\documentclass{article}

\usepackage[preprint]{neurips_2023}

\usepackage[utf8]{inputenc} 
\usepackage[T1]{fontenc}    
\usepackage{hyperref}       
\usepackage{url}            
\usepackage{booktabs}       
\usepackage{amsfonts}       
\usepackage{nicefrac}       
\usepackage{microtype}      
\usepackage{xcolor}         

\usepackage{subfiles}
\usepackage{csquotes}
\usepackage{lipsum}
\usepackage{bm}
\usepackage{amsmath}
\usepackage[noabbrev,capitalise]{cleveref}
\usepackage{graphicx}
\usepackage{amsthm}
\usepackage{enumitem}
\usepackage{amssymb}
\usepackage{xcolor}
\usepackage{colortbl}
\usepackage{wrapfig}
\usepackage{mathtools}
\usepackage{caption}
\usepackage{tikz}
\usetikzlibrary{calc,positioning,arrows}
\graphicspath{{./figures/}}
\usepackage{adjustbox}
\usepackage{multicol}
\usepackage{subcaption}

\newtheorem{example}{Example}

\newtheorem{definition}{Definition}

\newtheorem{claim}{Desideratum}
\newtheorem{property}{Property}

\crefname{claim}{Desideratum}{Desiderata}
\crefname{property}{Property}{Properties}

\newcommand{\name}{ICECREAM }      
\newcommand{\namenos}{ICECREAM}  
\newcommand{\namemeaning}{Identifying Coalition-based Explanations for Common and Rare Events in Any Model }       
\newcommand{\namemeaningnos}{Identifying Coalition-based Explanations for Common and Rare Events in Any Model}  

\newcommand{\ES}{\mathcal{E}}	
\newcommand{\parents}[1]{\bm{P\!A}_{#1}}

\title{Beyond Single-Feature Importance with \namenos}

\author{%
  Michael Oesterle\thanks{This work was done while M. Oesterle was working at Amazon.} \\
  University of Mannheim, Germany \\
  \texttt{michael.oesterle@uni-mannheim.de} \\
  \And
  Patrick Bl{\"o}baum \\
  Amazon, T{\"u}bingen, Germany \\
  \texttt{bloebp@amazon.com} \\
   \And
  Atalanti A. Mastakouri \\
  Amazon, T{\"u}bingen, Germany \\
  \texttt{atalanti@amazon.com} \\
  \And
  Elke Kirschbaum \\
  Amazon, T{\"u}bingen, Germany \\
  \texttt{elkeki@amazon.com} \\
}

\begin{document}

\maketitle

\begin{abstract}
    \emph{Which set of features was responsible for a certain output of a machine learning model? Which components caused the failure of a cloud computing application?} These are just two examples of questions we are addressing in this work by \emph{\namemeaning} (\namenos). Specifically, we propose an information-theoretic quantitative measure for the influence of a \emph{coalition of variables} on the distribution of a target variable. This allows us to identify which \emph{set} of factors is essential to obtain a certain outcome, as opposed to well-established explainability and causal contribution analysis methods which can assign contributions only to \emph{individual} factors and rank them by their importance. In experiments with synthetic and real-world data, we show that \name outperforms state-of-the-art methods for explainability and root cause analysis, and achieves impressive accuracy in both tasks. 
\end{abstract}

\setcounter{footnote}{0} 

\section{Introduction} \label{sec:introduction}
Raising the question \emph{why something happened} is a fundamental aspect of human nature and has always been a major driver for scientific progress across all disciplines. Moreover, answering this question is crucial to identify and resolve issues in complex systems. For instance, on-call engineers of a cloud computing application need to find out why their application is experiencing increased error rates to understand how to fix the problem and restore normal operations. 

In recent years, the question \enquote{Why does the machine learning model return a certain result for a given input?} has become particularly important with the tremendous advancements of machine learning (ML) and artificial intelligence (AI) and their application in various disciplines, from life sciences and medicine all the way to political decision making \citep[e.g.][]{emmert-2020,zhang-2022}. Understanding how black-box models work and knowing which input features most influence the model output has become a core goal in the growing field of explainability in AI (XAI) \citep[e.g.][]{lundberg-2017,burkart-2020,gerlings-2021}. Answering this question, however, is extremely challenging since the inputs to ML models can easily be composed of hundreds or even hundreds of thousands of input features (e.g., pixels in computer vision (CV)), and the best-performing models are usually highly complex deep neural networks. Various methods have been developed to rank the importance of individual input features for the model output, ranging from gradient-based approaches \citep[e.g.][]{sundararajan-2017} to methods using Shapley values \citep[e.g.][]{lundberg-2017,fryer-2021} to counterfactual reasoning \citep[e.g.][]{keane-2020,vonkuegelgen-2022}. In many applications, however, the explanation for an event is not just the behavior of a \emph{single} factor, but the combination of multiple factors that \emph{only jointly} result in the observed outcome. 

Identifying which coalition of factors is responsible for a given outcome provides valuable insights into the system under consideration. One such insight, for instance, is understanding which input factors need to be changed to obtain a desired change in the output.
Let us take a look at two simple examples where not a \emph{single} factor, but a coalition of variables is important for a certain outcome:

\begin{example} \label{ex:motivation}
    Consider a cloud computing application that is set up with redundancies to prevent errors from propagating through the system. The system only fails when multiple components experience an issue. 
    A simple example of such a system are two binary input variables $X_1$ and $X_2$ and a binary target $Y$. The input variables independently follow Bernoulli distributions with parameters $p_1$ and $p_2$, respectively, and the target value $y$ is the deterministic \emph{AND} of the inputs: $Y = X_1 \wedge X_2$. The corresponding graphical model is shown in \cref{fig:example} and all possible outcomes for this setup can be found in \cref{tab:example-outcomes}.
    As long as at least one of the two input factors is \enquote{healthy} (i.e., $X_1=0$ or $X_2=0$), the target does not experience an issue. However, if both inputs encounter an error (i.e., $X_1=X_2=1$), this leads to a failure in the target $Y$. 
\end{example}
 
\begin{table}[t]
\centering
	\begin{minipage}{.5\textwidth}
        \includegraphics[width=\textwidth]{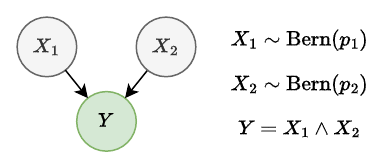}
        \captionof{figure}{Graphical causal model for \cref{ex:motivation}. The variables $X_1$ and $X_2$ independently follow Bernoulli distributions, while $Y$ is completely determined by the values of its causal parents. \label{fig:example}}    
    \end{minipage}
\hfill
	\begin{minipage}{.45\textwidth}
		\centering
		\caption{All possible outcomes for \cref{ex:motivation}. Additionally, we provide the explanation scores from \cref{eq:explanation-score-kl} in \cref{sec:explanation} for parameter values $p_1=0.1$, $p_2=0.8$. \label{tab:example-outcomes}}
		\begin{adjustbox}{max width = \textwidth}
		\begin{tabular}{rcccc}
		\toprule
		&case 1: &case 2: & case 3: & case 4: \\
		\midrule
		$x_1$ & 0 & 0 & 1 & 1\\
		$x_2$ & 0 & 1 & 0 & 1 \\
		$y$ & 0 & 0 & 0 & 1 \\
		\midrule
		$\ES(\{X_1\})$ & $1.0$ & $1.0$ & $-18.3$ & $0.9$ \\
		$\ES(\{X_2\})$ & $1.0$ & $-0.3$ & $1.0$ & $0.1$ \\
		$\ES(\{X_1,X_2\})$ & $1.0$ & $1.0$ & $1.0$ & $1.0$ \\
		\bottomrule
		\end{tabular}
		\end{adjustbox}
	\end{minipage}
\end{table}

\begin{example} \label{ex:credit}
    Consider an ML model that predicts the risk of credit applications based on the South German Credit dataset which contains various input features like age, housing situation, credit amount, duration, purpose, etc.\ \citep{gromping-2019}. For a specific application, the model predicts a high credit risk, and we would like to know why the model came to this conclusion and which features were relevant for the decision. 
    We can use e.g.\ the well-established SHAP method \citep{lundberg-2017} to quantify the importance of the individual features for this particular prediction. This ranking shows that the credit purpose, housing situation, and personal status are the 3 most important features for this output.\footnote{This is the result for a particular trained model and can, of course, vary depending on the used ML model.} However, it does not tell us what combination of features was crucial for the outcome, and how many features are actually required to explain the target value. One possibility would be to consider a fixed number of features or or to merge them into high-level features. But as we show e.g.\ in \cref{app:merging} this does not resolve the problem.
\end{example}

In this paper, we present \emph{\namenos}, a novel approach for \emph{\namemeaningnos}. \name is designed to detect \emph{coalitions of variables} whose interplay explains the observed outcome, instead of ranking single features based on their individual importance. 
For this we define an \emph{explanation score} which uses an information-theoretic approach to measure the contribution of groups of variables, inspired by the work on optimal feature selection by \citet{koller-1996}. \name can be used for explainability and interpretability of any feature/target system. Moreover, by incorporating a causal notion, it can also be applied in any system with a known causal structure, e.g., to perform root cause analysis (RCA) in a cloud computing system. 
We provide the required background on graphical causal models in \cref{sec:background}, introduce the proposed method in \cref{sec:explanation}, discuss related work in \cref{sec:related-work}, and present experiments with both synthetic and real-world data in \cref{sec:experiments}. Finally, we conclude the paper in \cref{sec:conclusion}.


\section{Graphical causal models} \label{sec:background}

Many systems for which we want to provide explanations can be represented as \emph{graphical causal models} \citep{pearl-2009}. For instance, \cref{fig:example} shows the graphical causal model corresponding to \cref{ex:motivation}: three observed variables where $X_1$ and $X_2$ causally influence $Y$. In general, these models represent the causal relations in a system via a directed graph $G=(\bm{V}, \bm{E})$ with nodes $\bm{V}=\{V_1, ..., V_N\}$ corresponding to random variables (RVs) $V_i$ with domains $\mathcal{V}_i$. An edge $(V_i, V_j) \in \bm{E}$, also represented in the graph as $V_i \rightarrow V_j$, means that $V_i$ has a direct causal influence on $V_j$. 

When a variable $V_i$ is set to some value $v_i$ in a hypothetical scenario where the other variables keep their relationships intact, this is called a (hard) \emph{intervention} on $V_i$. In the graph, such an intervention is represented by cutting all edges between $V_i$ and its causal parents $\parents{i}$. 
Pearl's \emph{do-operator} formalizes the change of the joint distribution $\mathbb{P}[V_1,\dots,V_N]$ that results from such an intervention and provides ways to compute the interventional distribution $\mathbb{P}[V_j \mid do(V_i=v_i)]$ under certain conditions \citep[Chapter 3.4]{pearl-2009}. We use the shorter notation $\mathbb{P}[V_j \mid do(V_i)]$ whenever the value $v_i$ of the intervention can be inferred from context.

Note that this interventional distribution must not be confused with the observational conditional distribution $\mathbb{P}[V_j\mid V_i=v_i]$: For example, conditional probabilities do not distinguish between the causal and anti-causal directions, while interventional probabilities do. The reason is that knowing that an effect happened also makes us update the probability of its causes, while intervening on an effect does not affect its causes at all. The interventional and conditional distributions only coincide if the variable $V_i$ has no causal parents.

We assume that the system under consideration can be partitioned into two sets of variables: first, a set $\bm{X}=\{X_1,\dots,X_n\}$ of observed (endogenous) variables; and second, a set of unobserved (exogenous) noise variables $\bm{\Lambda}=\{\Lambda_1,\dots,\Lambda_m\}$ which are jointly statistically independent, with $\bm{V}=\bm{X}\cup\bm{\Lambda}$ and $m\geq n$ such that each observed variable $X_i$ is causally influenced by at least one noise variable $\Lambda_i$. Without loss of generality, we assume that the observed variables $\bm{X}$ are in topological order in the graph $G$, i.e., $i < j$ for all $(X_i,X_j)\in\bm{E}$. 
For the remainder of this paper, we consider $X_n$ to be the \emph{target} whose value we want to explain, and denote this variable with the alias $Y$ to distinguish it from the other variables in the system. We hereafter denote the remaining set of observed variables with $\bm{X} := \{X_1, \dots, X_{n-1}\}$. 
Further we will consider \emph{coalitions of variables}, i.e., subsets of $\bm{V}$, denoted by $\bm{V}_C := \{V_i: i \in C\}$ with $C \subseteq I := \{1, \dots, N-1\}$. These coalitions can contain observed as well as unobserved variables from the graph. 
Note that we assume knowledge of the causal structure between the target $Y$ and all variables that are considered as possible explanations for the value of $Y$. However, this does not mean that we require the full causal graph of the system. For instance, if we consider the observed variables $\bm{X}$ as possible explanations of $Y$, we only need to know the causal structure among these variables and $Y$ while hidden confounders among the variables $\bm{X}$ are allowed. Admittedly, this is still a strong restriction in general, as discovering a causal graph from data is challenging and the subject of ongoing research \citep[e.g.][]{spirtes-1993,peters-2017}. Nevertheless, as we will discuss in \cref{sec:experiments}, for certain applications it is fair to assume the relevant causal structures to be known. Particularly, in the XAI setting the relevant graph structure is a trivial one-layer graph \citep[cf.][]{janzing-2020}.
Further, we assume that only interventions are possible that have a positive joint observational probability (in analogy to the positivity assumption in the potential outcomes framework \citep{rubin-2005}).


\section{\namenos: \namemeaning} \label{sec:explanation}
In this section, we present \namenos, a novel approach to find explanations for both common and extreme events. In contrast to established approaches for explainability and causal contribution analysis, \name is designed to handle \emph{coalitions} of variables instead of ranking the importance of \emph{individual} features. To achieve this, we first propose and analyze an \emph{explanation score} which quantifies the influence of a set of variables on the target variable. Then we discuss how this score can be applied in practice to address the following question:
\emph{What is the smallest coalition of variables which fully explains the target value $y$ for an observation $\bm{v}$?}

Obviously, the set of all variables provides a full explanation for any target value in a given system. However, in order to obtain actionable insights for the system and inspired by \emph{Occam's Razor} as well as the fact that humans understand simple explanations better than complex ones, we are interested in a concise and actionable explanation, i.e., one that includes as few variables as possible. 

First, we want to mention some properties we would like the explanation score used in \name to satisfy, and which are hence taken as constraints in the design of the score: 

\begin{claim} \label{cl:addition}
    Explanation scores of coalitions of variables should not be just the sum of the explanation scores for the individual variables. 
\end{claim}
Consider the case $x_1=x_2=y=0$ in \cref{ex:motivation}. Here, $x_1=0$ alone already fully explains the target value $y=0$, and the same is true for $x_2=0$. Hence the coalition $\{X_1, X_2\}$ should not have a higher explanation score than the two variables individually (as would be the case for additive scores). 

\begin{claim}\label{cl:rarity}
	Rare events which are required for the considered outcome should get a higher explanation score than common events.
\end{claim}
Consider again \cref{ex:motivation} and assume that $p_1=0.1$ and $p_2=0.8$. In this case, $X_1$ would be rarely 1, while $X_2$ is 1 most of the time. Intuitively---and as argued in \citet{budhathoki-2022}---, we consider $X_1$ to be the more interesting explanation for the outcome $y=1$. We expect this to also be reflected in the explanation score by assigning a higher score to $\{X_1\}$ than to $\{X_2\}$ in this example.  

\begin{claim}\label{cl:causal}
	Explanation scores in anti-causal direction should be zero.
\end{claim}
We consider this an important property as in a causal perspective it would not make sense for the value of an effect to explain its causes. In \cref{ex:motivation}, for instance, the explanation score of $\{Y\}$ for any value of $X_1$ or $X_2$ should always be zero since $Y$ cannot influence $X_1$ and $X_2$. 

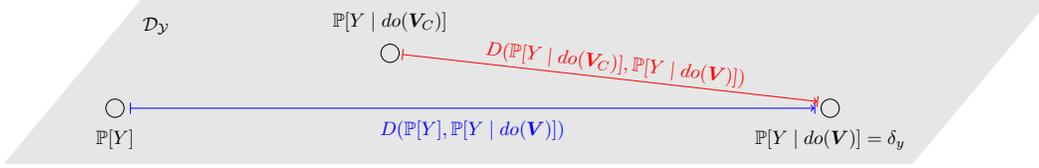
\begin{figure}[t]
\centering
\adjustbox{max width=\textwidth}{

\begin{tikzpicture}
	\draw[-, fill=black, draw=black, opacity=.1] (-2,-1) -- (14.5,-1) -- (17,2) -- (0.5,2) -- (-2,-1);
	\node at (0.75,1.5) {$\mathcal{D}_{\mathcal{Y}}$};
	
	\node [draw, circle] at (0,0) (py) {};
	\node[below = 0.1cm of py] {$\mathbb{P}[Y]$};
	
	\node [draw, circle] at (5,1) (pdo) {};
	\node[above = 0.1cm of pdo] {$\mathbb{P}[Y\mid do(\bm{V}_C)]$};
	
	\node [draw, circle] at (13,0) (pfull) {};
	\node[below = 0.1cm of pfull] {$\mathbb{P}[Y\mid do(\bm{V})]=\delta_y$};
	
	\draw[|->|, color=blue] ($(py.east)+(0.1,0.0)$) -- ($(pfull.west)+(-0.1,0.0)$);
	\node[color=blue] at (6.5, -.4) {$D(\mathbb{P}[Y], \mathbb{P}[Y\mid do(\bm{V})])$};
	
	\draw[|->|, color=red] ($(pdo.south east)+(0.1,0.1)$) -- ($(pfull.north west)+(-0.1,0.0)$);
	\node[color=red, rotate=-6.5] at ($(pdo)!0.5!(pfull)+(0.1,0.3)$) {$D(\mathbb{P}[Y\mid do(\bm{V}_C)], \mathbb{P}[Y\mid do(\bm{V})])$};
	
\end{tikzpicture}}
     \caption{Distances considered for the explanation score. In the space $\mathcal{D}_{\mathcal{Y}}$ of all distributions on the domain $\mathcal{Y}$, we consider (1) the distribution $\mathbb{P}[Y\mid do(\bm{V})]$ where all variables in the system have been fixed to some values $\bm{v}$, (2) the distribution $\mathbb{P}[Y\mid do(\bm{V}_C)]$ where only the values of the coalition $\bm{V}_C$ have been fixed, and (3) the distribution $\mathbb{P}[Y]$ where no values are fixed. If fixing the coalition values reduces the distance to the point mass distribution $\delta_y$ (red distance) compared to fixing nothing (blue distance), we conclude that the coalition provides at least a partial explanation for the value $Y=y$. In this case the explanation score is $\ES_{\bm{v}}(\bm{V}_C)>0$. If the distribution after fixing the coalition even equals the point mass distribution, the coalition fully explains the value $y$ and gets an explanation score of $\ES_{\bm{v}}(\bm{V}_C)=1$. On the other hand, if fixing the coalition values increases the distance to the point mass distribution compared to fixing nothing, then the coalition is not explaining the value of $Y$ at all, and gets a negative explanation score $\ES_{\bm{v}}(\bm{V}_C)<0$.} \label{fig:grounding}
\end{figure}

To obtain an explanation score that satisfies \cref{cl:addition,cl:causal,cl:rarity}, we propose to derive the score from the change in the probability distribution of the target variable $Y$. Specifically, we consider the following three distributions: (1) the distribution $\mathbb{P}[Y\mid do(\bm{V})]$ where \emph{all} variables in the system have been set to a fixed value; (2) the distribution $\mathbb{P}[Y\mid do(\bm{V}_C)]$ where only the values of the considered coalition of variables $\bm{V}_C$ have been fixed; and (3) the distribution $\mathbb{P}[Y]$ where no values are fixed at all. We then measure the distance between the distribution where the coalition values are fixed and the distribution where all values are fixed. If these two distributions are very close to each other, we conclude that the coalition $\bm{V}_C$ already determines a huge part of the behavior of $Y$. However, just considering the absolute distance between these two distributions does not provide the insights we are looking for -- unless in the special case where the distributions are equal and we can conclude that the coalition $\bm{V}_C$ fully explains the target value. For this reason we also consider the distance between the distribution with all values fixed to the distribution where no values are fixed. This way we know \emph{how much closer} fixing the coalition values gets us to the distribution with all values fixed compared to fixing nothing.
A sketch of the intuition behind the score is shown in \cref{fig:grounding}.

\begin{definition}[Explanation score] \label{def:explanation-score}
    Let $\mathcal{D}_{\mathcal{Y}}$ denote the set of all probability distributions over the domain $\mathcal{Y}$. Further let $D: \mathcal{D}_{\mathcal{Y}}^2 \rightarrow \mathbb{R}$ be a statistical distance measure to quantify the distance between distributions in $\mathcal{D}_{\mathcal{Y}}$. 
  
    The \emph{explanation score} of the coalition $\bm{V}_C \subseteq \bm{V}$ with respect to the observation $\bm{v} \in \bm{\mathcal{V}}$ is then defined as the function $\ES_{\bm{v}}: 2^{I} \rightarrow (-\infty, 1]$, with
    \begin{align}\label{eq:explanation-score}
        \ES_{\bm{v}}(\bm{V}_C) = 1 - \frac{D(\mathbb{P}[Y\mid do(\bm{V}_C)], \mathbb{P}[Y\mid do(\bm{V})])}{D(\mathbb{P}[Y], \mathbb{P}[Y\mid do(\bm{V})])} \; .
    \end{align}
\end{definition}

Note that the distance measure does not have to be a metric on $\mathcal{D}_{\mathcal{Y}}$, as long as it satisfies non-negativity (i.e., $D(P, Q) \geq 0 \; \forall P, Q \in \mathcal{D}_{\mathcal{Y}}$) and identity of indiscernibles (i.e., $D(P, Q) = 0 \, \Leftrightarrow \, P=Q$). Further note that the explanation score is only defined if $\mathbb{P}[Y] \neq \mathbb{P}[Y \mid do(\bm{V})]$, which intuitively makes sense as a zero-entropy distribution does not require any explanation for its value.

\Cref{tab:example-outcomes} provides the explanation scores for \cref{ex:motivation}, showing that this score satisfies \cref{cl:addition,cl:rarity}. 
Our approach is inspired by the work on optimal feature selection by \citet{koller-1996}, but differs from it in two main aspects: (1) We \emph{intervene} on a set of variables rather than conditioning on it. This way our explanation score distinguishes causal and anti-causal explanations as described in \cref{cl:causal}. (2) We compare the distance between intervening on a coalition vs.\ intervening on all variables to the distance between no intervention at all vs.\ intervening on all variables. As a consequence, the explanation score in \cref{def:explanation-score} is comparable across observations and can be shown to satisfy the following properties (see \cref{app:proofs} for the formal proofs):
\begin{property}\label{p:esvalues}
	 A coalition $\bm{V}_C$ gets an explanation score $\ES_{\bm{v}}(\bm{V}_C)> 0$ if and only if fixing the coalition values brings the distribution of the target variable closer to the point mass distribution $Y \sim \delta_{y}$.\footnote{The distribution of a RV $R$ is called a \emph{point mass distribution} (written as $R \sim \delta_r$) if $\exists r: \; \mathbb{P}[R=r] = 1$. Note that the distribution $\mathbb{P}[Y\mid do(\bm{V})]$, where all variables in the system have been fixed, is always a point mass distribution $Y\sim\delta_{y}$ as all noise variables $\bm{\Lambda}\subseteq\bm{V}$ have been fixed as well in this scenario.} 
	$\ES_{\bm{v}}(\bm{V}_C) = 1$ if and only if $\mathbb{P}[Y\mid do(\bm{V}_C)]=\mathbb{P}[Y\mid do(\bm{V})]=\delta_y$. In this case we say the coalition \emph{fully explains} the target value. 
	Conversely, $\bm{V}_C$ gets an explanation score $\ES_{\bm{v}}(\bm{V}_C)< 0$ if and only if fixing the coalition values moves the distribution of the target variable further away from the distribution $Y\sim\delta_y$, and it is $\ES_{\bm{v}}(\bm{V}_C)=0$ if and only if fixing the coalition values has no effect on the distance between the distribution of $Y$ and $\delta_y$.
\end{property}

\begin{property}\label{p:irrelevance}
    If a variable $V_i\in\bm{V}$ is irrelevant for $Y$, i.e., $\mathbb{P}[Y\mid do(V_i=v_i)]=\mathbb{P}[Y\mid do(V_i=v_i')]$ for all $v_i,v_i'\in\mathcal{V}_i$, then $\ES_{\bm{v}}(\bm{V}_C) = \ES_{\bm{v}}(\bm{V}_C \cup \{V_i\}) \; \forall \bm{v} \in \bm{\mathcal{V}}, \bm{V}_C \subseteq \bm{V} \setminus \{V_i\}$.
\end{property}

\begin{property} \label{p:monotonicity}
    If $\ES_{\bm{v}}(\bm{V}_C)=1$ for some $\bm{V}_C \subseteq \bm{V}$, then $\ES_{\bm{v}}(\bm{V}_{C'})=1$ for all $C' \supseteq C$.
\end{property}

\begin{property}\label{p:intervention}
	If $\ES_{\bm{v}}(\bm{V}_C)=1$ for some $\bm{V}_C \subseteq \bm{V}$, then there exists no $V_i\in \bm{V}\setminus\bm{V}_C$ such that changing the value of $V_i$ changes the value of the target $Y$. Further, to change the target value from $y$ to $y^*$, at least a subset $\bm{V}_S \subseteq \bm{V}_C, \bm{V}_S \neq \varnothing$ needs to change its values from $\bm{v}_S$ to some $\bm{v}^*_S$. 
\end{property}

In practical applications and for the remainder of this work, we use the KL divergence (i.e., the cross-entropy) between probability distributions as the distance measure for the explanation score: $D(P, Q) := D_{KL}(Q || P)$ (note the argument swap). With the KL divergence and $\mathbb{P}[Y\mid do(\bm{V})]=\delta_y$, the explanation score simplifies to 
\begin{align}\label{eq:explanation-score-kl}
  \ES_{\bm{v}}(\bm{V}_C) = 1 - \frac{\log{\mathbb{P}[Y=y\mid do(\bm{V}_C)]}}{\log{\mathbb{P}[Y=y]}} \; .
\end{align}

To identify the minimal coalitions that explain the target value, we loop through all possible coalitions of variables $\bm{V}_C$, ordered by increasing size, and calculate the explanation score for each coalition. For any coalition size $k$, we check if there is at least one coalition whose explanation score reaches a given threshold $\alpha \in [0, 1]$.\footnote{Note, to find coalitions fully explaining the target value it should be $\alpha=1$. In practice we choose slightly lower thresholds and accept \enquote{good} explanations by small coalitions instead of full explanations by large ones.} If so, we return all coalitions of this size with $\mathcal{E}(C) \geq \alpha$. Otherwise, we proceed to coalitions of size $k+1$.\footnote{Since $\ES_{\bm{v}}(\bm{V}) = 1$ for any observation $\bm{v}$, this procedure is guaranteed to terminate.} Note that this procedure will not return \emph{all} coalitions explaining the target value, but the set of the \emph{smallest} coalitions explaining the target value by the desired degree. 
With a minimum-size coalition with explanation score equal to one (which we call a \emph{minimal full explanation}), we can trace back the observed target value to a minimal set of variables that produced this target value; thus we obtain the most concise causal explanation possible for the observation.

In many situations we not only want to know why something happened, but also what we need to do to obtain another (desired) outcome. We can also leverage \name and its explanation score for this related task: From \cref{p:intervention} we know that, for a coalition with explanation score $\ES_{\bm{v}}(\bm{V}_C)=1$, it only makes sense to intervene on (a subset of) the variables of the coalition $\bm{V}_C$ to change the outcome. Hence we can start the search for an optimal intervention by looping through all possible values of the variables in the coalition $\bm{V}_C$ and identifying the one that maximizes the explanation score for a desired target value $y^*$. This approach is analogous to the \emph{directive explanations} recently proposed by \citet{singh-2023}. A naive optimization by looping over all explanation scores, however, becomes intractable very quickly with increasing number of variables and possible realizations thereof. Hence further research is required to develop an algorithm to handle real-world systems efficiently.


\section{Related work} \label{sec:related-work}
The idea of identifying a subset of relevant features using information-theoretic criteria was already explored in \citet{koller-1996} in the context of feature selection. The authors use the KL divergence between $\mathbb{P}[Y\mid\bm{V}]$ and $\mathbb{P}[Y\mid\bm{V}_C]$ as a measure for the loss of information when only considering the features $\bm{V}_C$. They then provide a backward elimination algorithm to find an approximately optimal feature set while staying close to the original target distribution. As a feature selection approach, this method averages over all samples, in contrast to \name which explains individual observations. Instance-wise feature selection methods \citep{shrikumar-2017,chen-2018,yoon-2018} and their extension \emph{instance-wise feature grouping} \citep{masoomi-2020} try to bridge this gap. However, they do not have a notion of causality, using mutual information to identify redundancies between features and labels instead. For \emph{causal} feature selection on time-series data, approaches using conditional independence testing have been proposed by \citet{mastakouri-2019,mastakouri-2021}. Further, \citet{wildberger-2023} proposed an interventional KL divergence to quantify the deviation between the distributions of ground truth and model under the same interventions. However, all these feature selection methods are not directly applicable in XAI. 

Feature relevance methods (see \citet{tritscher-2023} for a survey) express explanations as relevance (or \emph{contribution}) scores of individual features, measuring their respective importance for the model output. Here, it is generally accepted as a desired property that the contribution scores of all features add up to the difference between prediction and baseline -- called \emph{additivity}, \emph{completeness}, or \emph{efficiency}. This holds for all common feature relevance frameworks, including LIME \citep{ribeiro-2016}, integrated gradients \citep{sundararajan-2017}, and a variety of Shapley-based attribution methods \citep[e.g.][]{lundberg-2017,frye-2020,heskes-2020,wang-2020,janzing-2020,janzing-2021,jung-2022,bordt-2023}. 
Insisting on additivity, however, means that the interaction of feature coalitions cannot be adequately addressed by these methods (see \cref{cl:addition}).\footnote{A similar critique was already raised by \citet{grabisch-1999} for Shapley's original work on Shapley values for cooperative games \citep{shapley-1951}.} In \cref{app:merging} we show how coalitions of features can be considered in Shapley-based methods by merging them into a single, aggregated feature. This, however, does not solve the problem of additivity. Moreover, \citet{kumar-2020} pointed out that Shapley values do not satisfy human-centric goals of explainability, and \citet{jethani-2023} added that all the afore-mentioned methods can lead to false conclusions as they are \enquote{class-dependent}. As a remedy, they propose a new \enquote{distribution-aware} paradigm and two corresponding methods, SHAP-KL and FastSHAP-KL. Similar to \namenos, these methods make use of the target distribution under perturbations, though not under causal interventions. 
\citet{janzing-2020}, on the other hand, propose the use of interventions instead of conditioning to explain the causal influence of features, but by using Shapley-values they face the same additivity issue mentioned above.

Other quantitative measures of relevance which explicitly refer to causation were presented by \citet{pearl-1999,halpern-2015}. 
Although \name uses a very different way of quantifying the relevance of variables, it is still closely related to the \emph{probabilities of causation} (i.e., sufficiency and necessity) \citep{pearl-1999}: a full explanation score $\ES_{\bm{v}}(\bm{V}_C) = 1$ implies that $\bm{V}_C=\bm{v}_C$ is \emph{sufficient} for the target value, while $\ES_{\bm{v}}(\bm{V}_C) = -\infty$ implies that $\bm{V}_C \neq \bm{v}_C$ is \emph{necessary} for the target value. In addition, the notion of \emph{graded causation} \citep{halpern-2015} links back to our \cref{cl:rarity} that the base probabilities (or \emph{normality}, as \citeauthor{halpern-2015} call it) need to be taken into account for explanations.

Closest to our approach of considering combinations of features rather than individual features is the work of \citet{sani-2021}. They propose explanations based on causal structure learning using high-level features. Besides some difference in the assumptions made on the graph structure (e.g.\ they allow for hidden confounders between the features and the target, which we do not), the main difference lies in the applicable use cases: The approach in \citet{sani-2021} is ideal for explaining CV models where the image pixels can be summarized in interpretable features that are relevant for the model (e.g.\ in a bird classification task these could be \enquote{has red colored wing feathers} or \enquote{has a curved beak}). \namenos, on the other hand, is designed to identify coalitions explaining \textit{individual samples} instead of finding the high-level features that are \textit{generally} important for the model. Hence, \name can also be used to explain a specific prediction of a model (see \cref{sec:credit}) or to perform RCA (see \cref{sec:cloud}), which is not possible with the approach from \citet{sani-2021}.  


\section{Experiments and results} \label{sec:experiments}
We evaluate \namenos's performance in different use cases on synthetically generated and real-world datasets and benchmark it with state-of-the-art methods.
As a sanity check, we first apply \name to the CorrAL dataset \citep{john-1994}. This dataset was specifically designed to fool top-down feature selection methods. \name easily passes this test as all minimal full explanations consist exclusively of the relevant features according to the ground truth, whereas the irrelevant and correlated features are never picked (see \cref{app:corral} for details).

\subsection{Explaining an ML model on the South German Credit dataset} \label{sec:credit}
The South German Credit dataset \citep{gromping-2019} contains 1,000 loan applications with 21 features, and the corresponding risk assessment (low-risk or high-risk) as the binary target variable. 
We train a simple random forest classifier on the dataset whose outputs we want to explain.\footnote{Note that we want to explain this model and not the data itself. We can therefore ignore the fact that the classifier does not fit the data perfectly.}
We apply \name to the trained model, and compare the results with the instance-wise scores of the popular explainability framework SHAP \citep{lundberg-2017}. 

In order to obtain the interventional distributions required in \namenos, we follow the reasoning of \citet{janzing-2020} and consider the real world to be a confounder of the input features of an ML model, which therefore do not have direct causal links among each other. As a consequence, we can intervene directly on the input features by simply fixing the values of the features belonging to the considered coalition while sampling the remaining features from their (joint) marginal distribution.
Further, we choose a threshold of $\alpha = 0.9998$ to identify coalitions with (almost) full explanations while allowing for numerical errors. Additionally, we set the maximum coalition size to $k=4$ and stop the procedure if we do not find a coalition exceeding the threshold at this size. 
For SHAP we normalize the SHAP values $c(i), i \in I$ to $\sum{c(i)} = 1$ to get the relative importance of the features.

\begin{figure}[t]
    \centering
    \includegraphics[width=\textwidth]{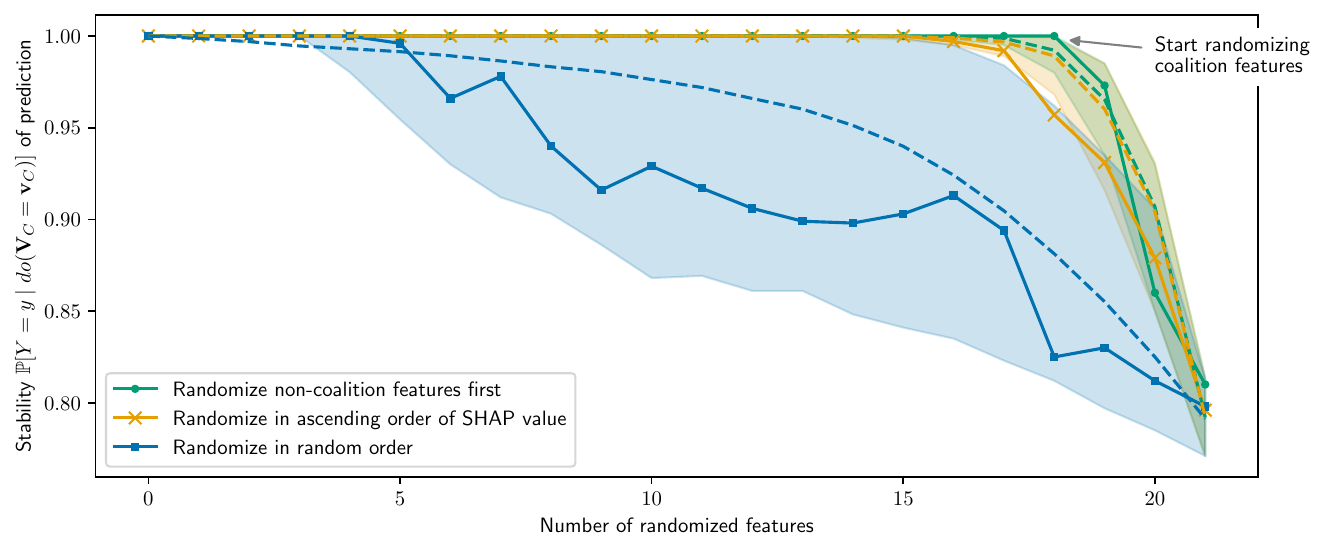}
    \caption{Stability of the prediction on the South German Credit dataset with respect to randomization of features. We randomize the values of the features in increasing order based on the SHAP values (blue), while ensuring that the features of the minimal coalition identified by \name are randomized last (green), and in completely random order (red). For the randomized features we pick 1,000 samples from their data distribution and check whether the model output changes. We do this for the sample mentioned in \cref{ex:credit} (solid lines) and over all samples for which a coalition with four or less components was identified (dashed line and colored area representing average and 95\% quantile). The model prediction remains stable as long as the features in the minimal coalition identified by \name remain fixed. When strictly following the SHAP value order, the prediction shows a very similar behavior. \label{fig:credit}}
\end{figure}

While \name returns the coalitions of features with explanation score exceeding the set threshold, SHAP has no notion of coalitions and provides a ranking of the different features given by their individual contribution scores. Hence the results are not directly comparable. 
Nevertheless, we can investigate whether the identified coalitions (for \namenos) and the set of top-$k$ features (for SHAP) are sufficient to determine the target label. 
For this we consider the samples for which we identify at least one coalition exceeding the set threshold for the explanation score. For each sample we randomize the input features in ascending order of the SHAP values for this sample, and create new predictions from the trained model (see \cref{fig:credit}, blue lines). Then we do the same, but ensure that the features of the coalitions identified by \name are randomized last (see \cref{fig:credit}, green lines). Additionally, as a baseline, we randomize the features in random order (see \cref{fig:credit}, red lines). If the identified features indeed fully explain the model output, the target values should be constant under the randomization of the remaining features. 

When considering for instance the sample from \cref{ex:credit}, \name identifies the coalition \{\texttt{age\_groups}, \texttt{housing}, \texttt{job\_status}\} to be a full explanation for the predicted high credit risk. As shown in \cref{fig:credit}, the target value is extremely stable as long as these coalition values are fixed even when randomizing over the remaining 18 features. In contrast to this, the top-3 features with respect to SHAP values are \texttt{credit\_purpose}, \texttt{housing}, and \texttt{personal\_status}. However, in \cref{fig:credit} we see that the target value changes earlier when not fixing the coalition features: the prediction remains stable for the randomization over 16 features when fixing the top SHAP values instead of over 18 features as long as the coalition features are fixed. \Cref{fig:credit} also shows that, on average across all samples, fixing the features with top-$k$ SHAP values and fixing the features in \namenos's coalitions leads to similarly stable model outputs. Both approaches lead to drastically more stability in the model output than randomizing over the features in completely random order. A key feature of \name in this context is that the minimum-size coalitions include information about the number of features needed for stable predictions, while the SHAP values only provide an order of relevance. Therefore, we cannot know from the SHAP values alone whether two, three or five features are required for a full explanation. Moreover, there can be multiple, different minimal coalitions independently explaining the target, which could not be expressed by a feature ranking.

\subsection{Explaining issues in a cloud computing application} \label{sec:cloud}
We consider a cloud computing application in which services call each other to produce a result which is returned to the user at the target service $Y$. Services can either be stand-alone (like a database or a random number generator) or call other services and transform their input to produce an output. When we observe an error at the target service $Y$, we want to know which service(s) caused this error. 
To imitate such a system, we synthetically generate data from a system with ten binary variables. Each variable represents a service with an intrinsic error rate $p_i \in [0, 1]$ and a threshold $t_i \in \{0, ..., |\parents{i}|\}$ which determines how many parent services must experience an error such that the error is propagated. 
The specific network and parameters that we are using for sample generation can be found in \cref{app:cloud}.
To obtain a representative set of observations with errors at the target service, we generate 10 million samples from the joint distribution and select the ones where $Y = 1$. The actual ground truth root causes of these errors are the noise variables $\Lambda_i$ with $\Lambda_i = 1$, but for our experiments we consider them as unobserved and only use observations from the variables $\bm{X}$. 

We compare \name to the root cause analysis (RCA) approaches presented by \citet{budhathoki-2022}. We use this method with different anomaly detection methods: an information theoretic outlier score (IT-RCA), and a score based on the mean deviation (Mean-RCA). Additionally, we evaluate both methods with a cumulative threshold (picking features in ascending SHAP order until the sum of their contribution scores reaches a threshold, denoted with -c) and an individual threshold (picking all features whose individual contribution value reaches the threshold, denoted with -i). Finally, we add a simple traversal algorithm as baseline where we identify anomalies and consider the first anomaly in the graph as the root cause of the issue (see \cref{app:cloud-traversal}).

For \name we consider the set of noise variables $\Lambda$ as the potential root causes and hence look for the minimal coalition $\bm{V}_C=\bm{\Lambda}_C$ explaining an error at the target service $Y$. This choice follows the argumentation in \citet{budhathoki-2022}. The advantage of intervening on the noise variables is that the connections among the observed variables remain unchanged. This way it is possible to identify the variable which shows not just anomalous behavior per se, but anomalous behavior conditioned on the behavior of its parents. This allows us to identify the actual root cause of an event, instead of just identifying the variable with the most abnormal behavior.
However, intervening on the unobserved noise variables is highly non-trivial in practice. Nevertheless, for the error propagation in this application we can estimate the conditional noise distribution $\mathbb{P}[\bm{\Lambda}\mid\bm{X}=\bm{x}]$ from the observations given the dependency structure of the application. With this we are able to compute an \emph{expected explanation score} over this noise distribution:
$\bar{\ES}_{\bm{v}}(\bm{\Lambda}_C) := \mathbb{E}_{\bm{\lambda} \sim \bm{\Lambda}\mid\bm{X}=\bm{x}}[\ES_{\bm{\lambda}}(\bm{\Lambda}_C)]$.

\begin{figure}[t]
	\centering
    \begin{subfigure}[t]{0.47\textwidth}
        \includegraphics[width=\textwidth]{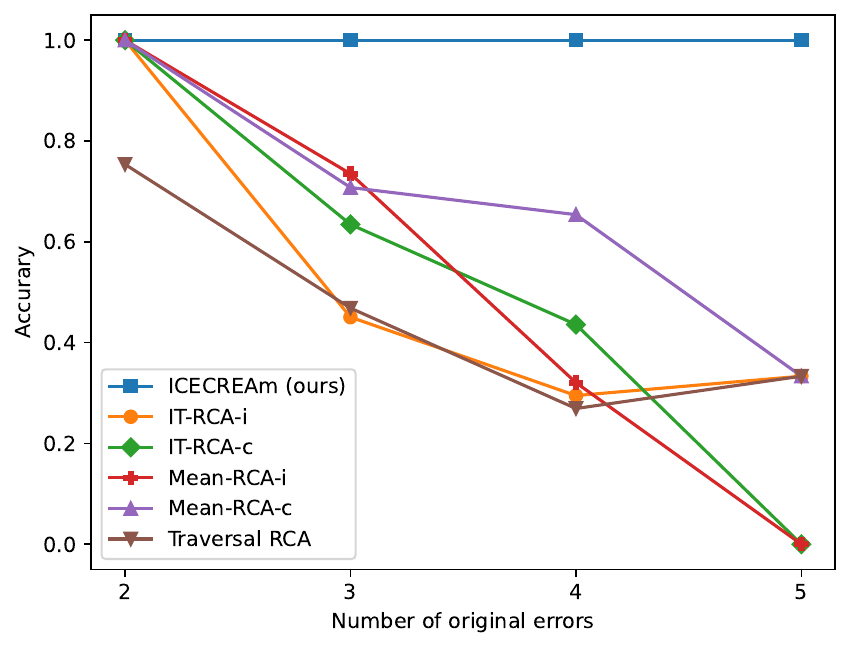}
        \caption{Accuracy of identifying root causes for different numbers of injected errors. \label{fig:cloud-accuracy}}    
    \end{subfigure}
    \hfill
    \begin{subfigure}[t]{0.47\textwidth}
        \includegraphics[width=\textwidth]{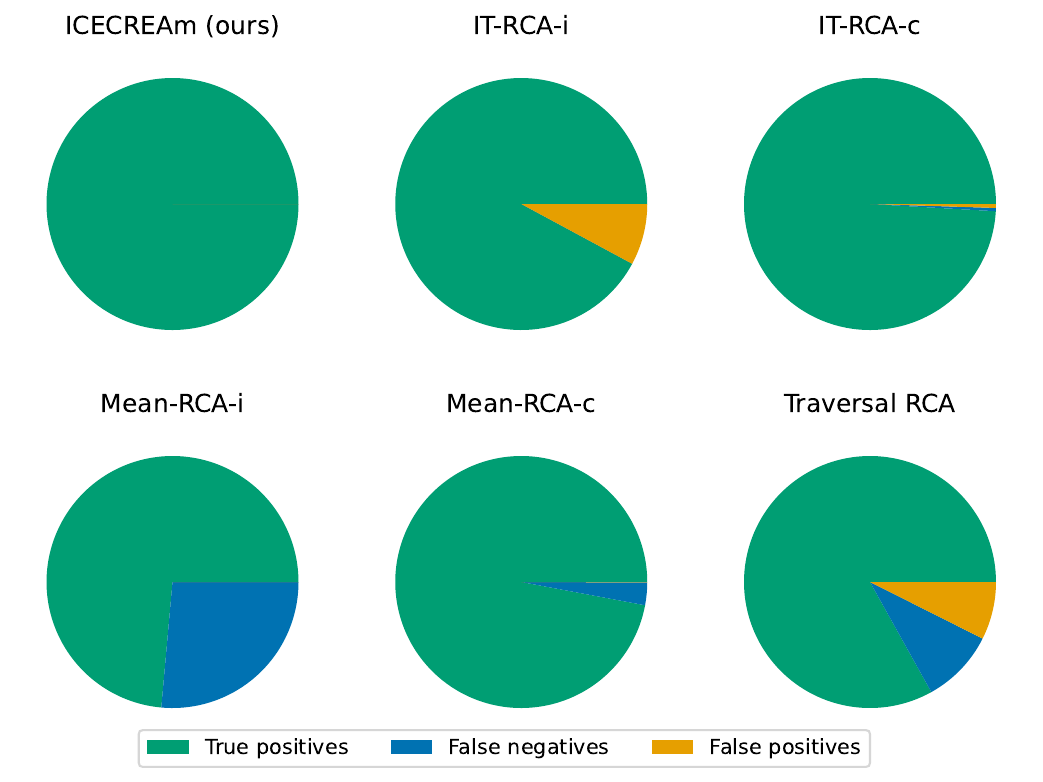}
         \caption{True positives, false positives, and false negatives for the cases with at least 3 injected errors. \label{fig:cloud-confusion}}
    \end{subfigure}	
	\caption{Accuracy for the identification of root causes of issues in a cloud computing application. If the ground truth root cause only consists of two errors in the system, most methods for RCA are able to identify them with 100\% accuracy. However, if the issue is the result of a combination of more than two errors, only \name is able to identify the coalition of failing services correctly, while the performance of the other RCA methods drops drastically.} \label{fig:cloud-results}
\end{figure}

We compare the methods with respect to accuracy, as well as true positives, false positives, and false negatives. For \namenos, the root cause of a sample counts as correctly identified if and only if all identified minimal coalitions are correct. For the other methods, the root cause counts as correctly identified if the returned set of root causes is correct.
\Cref{fig:cloud-results} shows the performance of the methods for different numbers of errors injected into the system. It is evident that \name is the only method which still achieves close to 100\% accuracy when the number of errors increases. 

\section{Conclusion} \label{sec:conclusion}
In this paper, we proposed a novel approach for explainability and causal contribution analysis which is able to identify minimal coalitions of variables that are required to explain a target value. 
We presented an explanation score which provides a quantitative, information-theoretic measure for the relative shift in the target distribution caused by an intervention on a coalition of variables, compared to the distribution before the intervention and the point mass distribution corresponding to the actual observation. 
This score satisfies all desired properties in order to explain target values which are the result of an interaction of multiple factors. Additionally, we proved some beneficial properties of the score which, for instance, allow for an easy interpretation of the score, and also allow us to use the explanation score to identify optimal interventions to change the target value. 

Our experiments show that \name achieves similar performance in explaining an ML model as the established SHAP values, while providing more concise information about coalitions instead of just a ranking of feature contributions. Moreover, it clearly outperforms state-of-the-art RCA methods in identifying the root cause of an issue, particularly when this is a combination of multiple factors.

The \name framework is sufficiently generic to be applied to any causal network; however, our current formulation relies on a few assumptions or design choices which limit its scope. For instance, the explanation score in its current form cannot be applied for \emph{continuous target variables}, since the point mass distribution $\delta_y$ is not a valid target distribution in this case. 
Additionally, for systems with many variables, iterating over all coalitions can become quite costly, particularly if no \enquote{small} coalition with \enquote{good} explanation score can be found and larger coalitions also need to be tested. This is the case despite the extremely fast estimation of the explanation score and simply caused by the combinatorial explosion for larger coalitions. In future work we will investigate how the fact that $\{C \in 2^I: \ES(C) = 1\}$ is an upper set of $2^I$ can be used to develop a more efficient search algorithm for minimal coalitions.

\bibliographystyle{apalike}
\bibliography{references.bib}

\newpage
\appendix

\setcounter{property}{0}

\section{Proofs} \label{app:proofs}
\begin{property}\label{p:esvalues}
    A coalition $\bm{V}_C$ gets an explanation score $\ES_{\bm{v}}(\bm{V}_C)> 0$ if and only if fixing the coalition values brings the distribution of the target variable closer to the point mass distribution $Y \sim \delta_{y}$.
    $\ES_{\bm{v}}(\bm{V}_C) = 1$ if and only if $\mathbb{P}[Y\mid do(\bm{V}_C)]=\mathbb{P}[Y\mid do(\bm{V})]=\delta_y$. In this case we say the coalition \emph{fully explains} the target value. 
    Conversely, $\bm{V}_C$ gets an explanation score $\ES_{\bm{v}}(\bm{V}_C)< 0$ if and only if fixing the coalition values moves the distribution of the target variable further away from the distribution $Y\sim\delta_y$, and it is $\ES_{\bm{v}}(\bm{V}_C)=0$ if and only if fixing the coalition values has no effect on the distance between the distribution of $Y$ and $\delta_y$.
\end{property}
\begin{proof}
    This directly follows from \cref{eq:explanation-score} in \cref{def:explanation-score}. We have
    \begin{align*}
        \ES_{\bm{v}}(\bm{V}_C) > 0 
        \quad&\Leftrightarrow\quad 1 - \frac{D(\mathbb{P}[Y\mid do(\bm{V}_C)], \mathbb{P}[Y\mid do(\bm{V})])}{D(\mathbb{P}[Y], \mathbb{P}[Y\mid do(\bm{V})])} > 0 \\
        \quad&\Leftrightarrow\quad \frac{D(\mathbb{P}[Y\mid do(\bm{V}_C)], \mathbb{P}[Y\mid do(\bm{V})])}{D(\mathbb{P}[Y], \mathbb{P}[Y\mid do(\bm{V})])} < 1 \\
        \quad&\Leftrightarrow\quad D(\mathbb{P}[Y\mid do(\bm{V}_C)], \mathbb{P}[Y\mid do(\bm{V})]) < D(\mathbb{P}[Y], \mathbb{P}[Y\mid do(\bm{V})]) \\
        \quad&\Leftrightarrow\quad \mathbb{P}[Y\mid do(\bm{V}_C)] \text{ is closer to } \mathbb{P}[Y\mid do(\bm{V})] \text{ than } \mathbb{P}[Y] \\
    \end{align*}
    \begin{align*}
        \ES_{\bm{v}}(\bm{V}_C) < 0 
        \quad&\Leftrightarrow\quad 1 - \frac{D(\mathbb{P}[Y\mid do(\bm{V}_C)], \mathbb{P}[Y\mid do(\bm{V})])}{D(\mathbb{P}[Y], \mathbb{P}[Y\mid do(\bm{V})])} < 0 \\
        \quad&\Leftrightarrow\quad \frac{D(\mathbb{P}[Y\mid do(\bm{V}_C)], \mathbb{P}[Y\mid do(\bm{V})])}{D(\mathbb{P}[Y], \mathbb{P}[Y\mid do(\bm{V})])} > 1 \\
        \quad&\Leftrightarrow\quad D(\mathbb{P}[Y\mid do(\bm{V}_C)], \mathbb{P}[Y\mid do(\bm{V})]) > D(\mathbb{P}[Y], \mathbb{P}[Y\mid do(\bm{V})]) \\
        \quad&\Leftrightarrow\quad \mathbb{P}[Y\mid do(\bm{V}_C)] \text{ is further away from } \mathbb{P}[Y\mid do(\bm{V})] \text{ than } \mathbb{P}[Y] \\
    \end{align*}
    \begin{align*}
        \ES_{\bm{v}}(\bm{V}_C) = 1 
        \quad&\Leftrightarrow\quad 1 - \frac{D(\mathbb{P}[Y\mid do(\bm{V}_C)], \mathbb{P}[Y\mid do(\bm{V})])}{D(\mathbb{P}[Y], \mathbb{P}[Y\mid do(\bm{V})])} = 1 \\
        \quad&\Leftrightarrow\quad \frac{D(\mathbb{P}[Y\mid do(\bm{V}_C)], \mathbb{P}[Y\mid do(\bm{V})])}{D(\mathbb{P}[Y], \mathbb{P}[Y\mid do(\bm{V})])} = 0 \\
        \quad&\Leftrightarrow\quad D(\mathbb{P}[Y\mid do(\bm{V}_C)], \mathbb{P}[Y\mid do(\bm{V})]) = 0 \\
        \quad&\Leftrightarrow\quad \mathbb{P}[Y\mid do(\bm{V}_C)] = \mathbb{P}[Y\mid do(\bm{V})] \\
    \end{align*}
    \begin{align*}
        \ES_{\bm{v}}(\bm{V}_C) = 0 
        \quad&\Leftrightarrow\quad 1 - \frac{D(\mathbb{P}[Y\mid do(\bm{V}_C)], \mathbb{P}[Y\mid do(\bm{V})])}{D(\mathbb{P}[Y], \mathbb{P}[Y\mid do(\bm{V})])} = 0 \\
        \quad&\Leftrightarrow\quad \frac{D(\mathbb{P}[Y\mid do(\bm{V}_C)], \mathbb{P}[Y\mid do(\bm{V})])}{D(\mathbb{P}[Y], \mathbb{P}[Y\mid do(\bm{V})])} = 1 \\
        \quad&\Leftrightarrow\quad D(\mathbb{P}[Y\mid do(\bm{V}_C)], \mathbb{P}[Y\mid do(\bm{V})]) = D(\mathbb{P}[Y], \mathbb{P}[Y\mid do(\bm{V})]) \\
        \quad&\Leftrightarrow\quad \mathbb{P}[Y\mid do(\bm{V}_C)] \text{ is as close to } \mathbb{P}[Y\mid do(\bm{V})] \text{ as } \mathbb{P}[Y]
    \end{align*}
\end{proof}

\begin{property}\label{p:irrelevance}
    If a variable $V_i\in\bm{V}$ is irrelevant for $Y$, i.e., $\mathbb{P}[Y\mid do(V_i=v_i)]=\mathbb{P}[Y\mid do(V_i=v_i')]$ for all $v_i,v_i'\in\mathcal{V}_i$, then $\ES_{\bm{v}}(\bm{V}_C) = \ES_{\bm{v}}(\bm{V}_C \cup \{V_i\}) \; \forall \bm{v} \in \bm{\mathcal{V}}, \bm{V}_C \subseteq \bm{V} \setminus \{V_i\}$.
\end{property}
\begin{proof}
  If $V_i$ is irrelevant for $Y$, the distribution of $Y$ does not change when we fix $V_i=v_i$, and therefore $\mathbb{P}[Y] = \mathbb{P}[Y\mid do(V_i)]$ as well as $\mathbb{P}[Y\mid do(\bm{V}_C)] = \mathbb{P}[Y\mid do(\bm{V}_C \cup \{V_i\})]$. 
\end{proof}

\begin{property} \label{p:monotonicity}
    If $\ES_{\bm{v}}(\bm{V}_C)=1$ for some $\bm{V}_C \subseteq \bm{V}$, then $\ES_{\bm{v}}(\bm{V}_{C'})=1$ for all $C' \supseteq C$.
\end{property}
\begin{proof}
  We assume without loss of generality that $\bm{V}_{C'} = \bm{V}_C \cup \{V_i\}$, $V_i\notin\bm{V}_C$. Then we have
  \begin{align*}
  \ES_{\bm{v}}(\bm{V}_C) = 1 \quad&{\Leftrightarrow}\quad
     D(\mathbb{P}[Y\mid do(\bm{V}_C)], \mathbb{P}[Y\mid do(V)]) = 0 \\
     & \Rightarrow \quad \mathbb{P}[Y\mid do(\bm{V}_C)] = \mathbb{P}[Y\mid do(V)] = \delta_y \\
   & {\Rightarrow}\quad  \mathbb{P}[Y=y\mid do(\bm{V}_C)] = 1\\
    &{\Rightarrow} \quad  \mathbb{P}[Y=y\mid do(\bm{V}_C), do(V_i=v_i) ] = 1  \quad \forall v_i\in\mathcal{V}_i \\
    	&\phantom{{\Rightarrow} \quad  \mathbb{P}[Y=y\mid do(\bm{V}_C), do(V_i=v_i) ] = 1   } \text{ with } \mathbb{P}[Y=y, \bm{V}_C=\bm{v}_C, V_i=v_i] > 0 \\ 
   & {\Rightarrow } \quad \mathbb{P}[Y=y\mid do(\bm{V}_{C'}) ] = 1 \\ 
   & {\Leftrightarrow } \quad \mathbb{P}[Y\mid do(\bm{V}_{C'})] = \mathbb{P}[Y\mid do(V)] \\
   &{\Leftrightarrow } \quad D(\mathbb{P}[Y\mid do(\bm{V}_{C'})], \mathbb{P}[Y\mid do(V)]) = 0  
\quad { \Leftrightarrow }\quad \ES_{\bm{v}}(\bm{V}_{C'}) = 1 \; .
  \end{align*}
\end{proof}

\begin{property}\label{p:intervention}
	If $\ES_{\bm{v}}(\bm{V}_C)=1$ for some $\bm{V}_C \subseteq \bm{V}$, then there exists no $V_i\in \bm{V}\setminus\bm{V}_C$ such that changing the value of $V_i$ changes the value of the target $Y$. Further, to change the target value from $y$ to $y^*$, at least a subset $\bm{V}_S \subseteq \bm{V}_C, \bm{V}_S \neq \varnothing$ needs to change its values from $\bm{v}_S$ to some $\bm{v}^*_S$. 
\end{property}
\begin{proof}
    For a coalition $\bm{V}_C$ with full explanation score, changing any non-coalition variables does not affect the target variable:
    \begin{align*}
        \ES_{\bm{v}}(\bm{V}_C) = 1 &\quad{\Leftrightarrow}\quad
            D(\mathbb{P}[Y\mid do(\bm{V}_C)], \mathbb{P}[Y\mid do(V)]) = 0 \\
        & \quad{\Rightarrow} \quad \mathbb{P}[Y=y\mid do(\bm{V}_C)] = 1 \\
        &\quad{\Rightarrow} \quad  \mathbb{P}[Y=y\mid do(\bm{V}_C), do(V_i=v_i) ] = 1  \quad \forall v_i\in\mathcal{V}_i, V_i\notin\bm{V}_C \\
                &\phantom{{\Rightarrow} \quad  \mathbb{P}[Y=y\mid do(\bm{V}_C), do(V_i=v_i) ] = 1   } \text{ with } \mathbb{P}[Y=y, \bm{V}_C=\bm{v}_C, V_i=v_i] > 0 \\ 
        &\quad{\Rightarrow}\quad   \mathbb{P}[Y=y^*\neq y\mid do(\bm{V}_C), do(V_i) ] = 0  \quad \forall v_i\in\mathcal{V}_i, V_i\notin\bm{V}_C
    \end{align*}
    From \cref{p:monotonicity} it follows that there exists a minimal coalition $\bm{V}_{min}\subseteq\bm{V}_C$ with full explanation score, which cannot be empty by construction of the explanation score (remember that $\ES(\bm{V}_\varnothing)$ is always $0$). If we remove another element $V_i$ from $\bm{V}_{min}$, the explanation score will be $<1$ and thus the probability for target values $y^*\neq y$ becomes non-zero.
\end{proof}

\section{Applying Shapley-based approaches to coalitions of variables} \label{app:merging}
In this section, we discuss how Shapley-based approaches can be applied to quantify the contribution of coalitions of variables.  

\begin{figure}[t]
\centering
    \begin{subfigure}{0.48\textwidth}
    \centering
        \includegraphics[width=\textwidth]{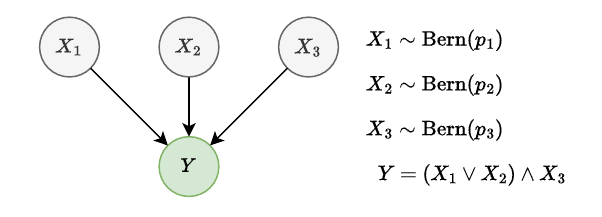}
        \caption{Original system \label{fig:merging-before}}
    \end{subfigure}
    \hfill
    \begin{subfigure}{0.48\textwidth}
    \centering
        \includegraphics[width=\textwidth]{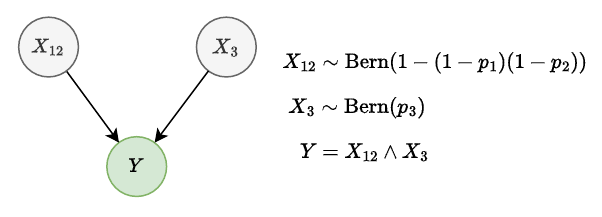}
        \caption{System after merging coalition variables \label{fig:merging-after}}
    \end{subfigure}	
	\caption{Illustration how Shapley-based approaches can be applied to coalitions of variables by merging the variables accordingly. In this example we consider the coalition of $X_1$ and $X_2$. We can merge these variables into a new variable $X_{12}=X_1\lor X_2$, and then compute the Shapley-based contribution of this variable in the new graph. Note that it is not always possible to merge variables like this, depending on the functional relationships in the network. But as a general approach we can allow for multi-dimensional variables as representations of the coalitions and then merge variables by creating a vector $X_{12} = (X_1, X_2)$. \label{fig:merging}}
\end{figure}

First, we consider a simple, intuitive approach that allows us to apply well-established Shapley-based contribution methods for individual variables \citep[e.g.][]{lundberg-2017,frye-2020,heskes-2020,wang-2020,janzing-2020,janzing-2021,jung-2022} to coalitions of variables.
The idea is to merge variables belonging to the respective coalition into a \enquote{super-variable} (which can be one- or multi-dimensional). For instance, in the example shown in \cref{fig:merging-before}, the variables $X_1$ and $X_2$ can be merged into a joint variable $X_{12} = X_1 \lor X_2$ to represent the coalition of the two variables (see \cref{fig:merging-after}). Then the Shapley-based attribution methods can simply be applied to the variable $X_{12}$ to obtain a contribution score for the coalition. 

The efficiency property of Shapley values ensures that any Shapley-based contributions of the individual variables in the original graph, denoted here by $\phi_{X_i}$, sum up to some statistic of the target variable
\begin{align}
\sum_{i\in I} \phi_{X_i} = \Phi_Y \, .
\end{align}

This statistic $\Phi_Y$ can, for instance, be the deviation from the expectation, an \emph{information-theoretic (IT) outlier score} of $Y$ \citep[see][]{budhathoki-2022}, or any other function which does not depend on the variables $X_i$, but only on the target distribution.

If we now merge the variables of a coalition $C \subset I$, we get a new graph, consisting of the \enquote{super-variable} $X_C$ as well as some (unchanged) variables $X_i, i \in I \setminus C$. Calculating the Shapley-based contribution values on this new set of variables can result in new values $\psi_{X_k}$ for all $X_k \in \{X_C\} \cup \{X_i:\, i\in I \setminus C\}$. But, since this merging of variables does, by definition, not affect the target variable or its distribution, the following still holds:

\begin{align}
    \psi_{X_C} + \sum_{i\in I\setminus C} \psi_{X_i} = \Phi_Y \, .
\end{align}
Consequently, it is
\begin{align}
    \sum_{i\in I} \phi_{X_i} = \psi_{X_C} + \sum_{i\in I\setminus C} \psi_{X_i}  \, .
\end{align}

Intuitively, we would expect that the contribution scores of the non-coalition variables should not depend on whether we merge the coalition variables or not, since their influence on the target has not changed. If we assume $\psi_{X_i}=\phi_{X_i}$ for all $i\in I\setminus C$, this necessarily results in the contribution of the coalition being the sum of the contributions of its components:
\begin{align}\label{eq:additive-con}
\psi_{X_C} = \sum_{i\in C} \phi_{X_i}\, .
\end{align}

To put it simply: Any Shapley-based contribution method either satisfies \cref{eq:additive-con} and hence violates \cref{cl:addition}, or allows the contribution of individual variables to change depending on whether and how the other variables are grouped into coalitions. We consider both options not desirable for a contribution score to quantify the influence of coalitions. 

Nevertheless, we consider examples for both options in our experiments with existing RCA methods in \cref{sec:cloud} for the comparison with \namenos. 
The first method uses the above-mentioned IT outlier score. This score leaves the contributions of the non-coalition nodes unchanged. As a result, merging variables just gives the same results as summing up the respective individual contribution values. 
The second approach uses a \emph{mean deviation outlier score}. This score changes the contributions of the non-coalition variables, and can therefore give a non-trivial coalition contribution.
Our experiments show that \name clearly outperforms both approaches in the identification of root causes in a cloud computing application for cases where multiple errors together result in a failure at the target service.

Only quite recently, more sophisticated approaches have been proposed to link Shapley values to coalitions of features, from the pairwise \emph{Shapley Interaction Values} \citep{lundberg-2020} to the \emph{Shapley-Taylor Interaction Index} \citep{dhamdhere-2020} and \emph{$n$-Shapley} \citep{bordt-2023}. These approaches make the connections between features more transparent and therefore interpretable, but they still follow the premise that contributions have to add up. If one coalition of features is very important, the remaining features have to be less important. What is measured is basically the importance of feature sets \emph{relative to each other}. Therefore, our above arguments with respect to \cref{cl:addition} still apply. In contrast, \name allows for independent explanation scores for all coalitions, which can be thought of describing the (causal) importance of a feature set relative to the natural target distribution and the actual observation.


\section{Experiments}
This section provides additional details on the experiments whose basic setup and results are reported in \cref{sec:experiments} in the main paper. 

\subsection{CorrAL feature selection dataset} \label{app:corral}
CorrAL is a synthetic dataset consisting of six Boolean features (\texttt{A0}, \texttt{A1}, \texttt{B0}, \texttt{B1}, \texttt{Irrelevant} and \texttt{Correlated}) and 160 samples. The feature names are rather self-explanatory as the target variable is defined as 
$$\texttt{class} := (\texttt{A0} \wedge \texttt{A1}) \vee (\texttt{B0} \wedge \texttt{B1})\; .$$
The remaining two features are not used---\texttt{Irrelevant} is uniformly random, and \texttt{Correlated} matches the target label in 75\% of all samples. \citet{john-1994}, who created the dataset as an illustration for the shortcomings of existing feature selection algorithms, report that greedy selection mechanisms (both top-down and bottom-up) have difficulties identifying the correct feature set. We therefore test \name on the full dataset and see which features are selected as part of the minimum-size coalitions with full explanation score.

As causal model for the dataset, we consider the input features to be independent, while we allow all of them to potentially influence the target. As threshold for coalitions with full explanation score we use $\alpha = 0.998$ and evaluate all coalitions sizes $k \in \{1,\dots, \text{\#features}\}$.

\name easily identifies the minimal full explanations in all cases without including the irrelevant or only correlated feature into the coalitions. As an example, let us consider the first row of the dataset, where all features are \texttt{False}. A full explanation must consist of at least one of \texttt{A0} and \texttt{A1} (to explain why the first conjunction is \texttt{False}), plus at least one of \texttt{B0} and \texttt{B1}. Accordingly, the minimum-size coalitions found by \name are the four combinations $\{\texttt{A0}, \texttt{B0}\}$, $\{\texttt{A0}, \texttt{B1}\}$, $\{\texttt{A1}, \texttt{B0}\}$, and $\{\texttt{A1}, \texttt{B1}\}$.

\subsection{Explaining an ML model on the South German Credit dataset} \label{app:credit}
The South German Credit (SGC) dataset \citep{gromping-2019} contains 1,000 loan applications with 21 features, and the corresponding risk assessment (low-risk or high-risk) as the binary target variable. 

\subsubsection{Model}
\begin{figure}[t]
    \centering
    \includegraphics[width=0.6\textwidth]{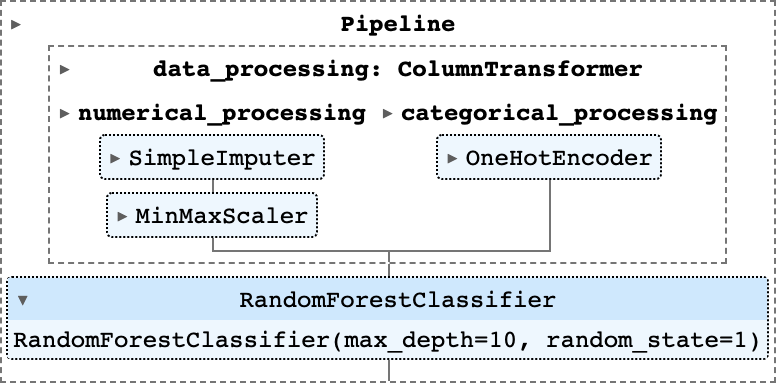}
    \caption{Training pipeline for the model used in the experiment on the SGC dataset. \label{fig:credit-pipeline}}
\end{figure}

In this experiment, we are explaining the predictions of a trained ML model on a given dataset. As the model, we use a simple random forest classifier with the pipeline shown in \cref{fig:credit-pipeline}, trained on the full dataset with a train/test split of 90/10. The fitted model reaches an accuracy of 96\% for the training set and 80\% for the test set. Note that we consider explanations for this model -- not for the data itself. Hence we ignore that the model is not performing extremely well. 

\subsubsection{SHAP values}
\begin{figure}[t]
    \centering
    \includegraphics[width=\textwidth]{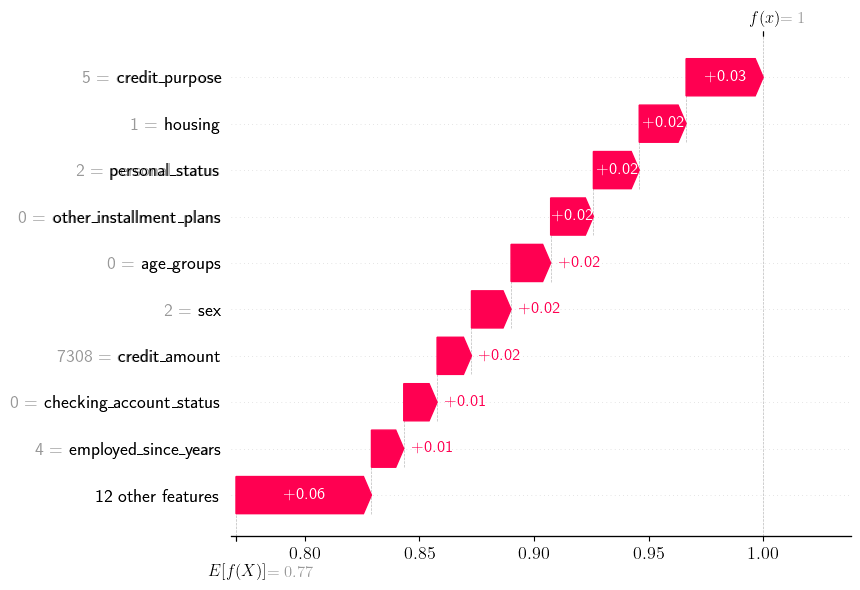}
    \caption{SHAP values for sample 500 of the SGC dataset. \label{fig:credit-shap}}
\end{figure}

To obtain the SHAP values for comparison, we make use of the \texttt{shap} package by \citet{lundberg-2017}. The use of their explainer is straight-forward; note that the categorical features need to be converted into integer categories first for \texttt{shap} to handle the input. In \cref{fig:credit-shap} we show the SHAP output for the sample with ID 500, the sample also shown in \cref{tab:sample-500} and mentioned in \cref{ex:credit} in the main paper.

\begin{table}[t]
\centering
\caption{Feature values and SHAP values for sample 500 of the SGC dataset. \label{tab:sample-500}}
    \begin{tabular}{llc}
        \toprule
        Feature & Value & SHAP value \\
        \midrule
        credit\_purpose & others & 0.034 \\
        housing & own & 0.020 \\
        personal\_status & married/widowed & 0.020 \\
        other\_installment\_plans & bank & 0.019 \\
        age\_groups & 0 & 0.017 \\
        sex & male & 0.017 \\
        credit\_amount & 7308 & 0.015 \\
        checking\_account\_status & ... < 0 DM & 0.015 \\
        employed\_since\_years & unemployed & 0.014 \\
        job\_status & mgmt/self-employed/highly qualified empl./officer & 0.014 \\
        property & real estate & 0.013 \\
        credit\_history & no credits taken/all credits paid back duly & 0.009 \\
        telephone & True & 0.007 \\
        num\_existing\_credits & 1 & 0.006 \\
        savings & unknown/no savings account & 0.004 \\
        credit\_duration\_months & 10 & 0.003 \\
        installment\_rate & 25 <= ... < 35 & 0.002 \\
        present\_residence\_since & >= 7 yrs & 0.001 \\
        other\_debtors\_guarantors & none & 0.001 \\
        foreign\_worker & False & 0.000 \\
        num\_people\_liable\_for & 0 to 2 & -0.002 \\
        \bottomrule
    \end{tabular}
\end{table}

Originally, SHAP values are designed to explain the difference between the actual prediction and its expected value. Since the ratio of low-risk ($\texttt{credit\_risk} = 0$) and high-risk ($\texttt{credit\_risk} = 1$) loan applications in the dataset is $\approx 30/70$, this difference is not always the same. We therefore normalize the SHAP values such that the feature contributions add up to 1 for each sample. This does not change the relative importance of the features, but allows for a better comparison across low-risk and high-risk samples.

\subsubsection{Comparison}
We compare the stability of the prediction under three different randomizations, showing the probability that the prediction remains unchanged for any number of randomized features. The underlying reasoning is: The better an explanation, the more stable the model prediction when the explanation holds (i.e., when the respective feature values are fixed). Therefore, we start with all features fixed to their observed values and successively randomize more and more features according to their joint distribution in the dataset, observing when the prediction stability drops. The three randomizations only differ in the \emph{order} of features:

\begin{description}
    \item[Randomize in ascending order of SHAP values] The features are ordered ascendingly by their SHAP value. This is the order in which the prediction can be expected to stay stable as long as possible: Randomizing features with a SHAP value $< 0$ means that the \emph{negative} influence of these features is removed. The features with the highest (positive) SHAP value, on the other hand, are kept fixed until much longer, such that their stabilizing influence can be used. 
    \item[Randomize non-coalition features first] The features of the minimum-size full explanation identified by \name are randomized \emph{last}. By doing so, we see whether the prediction is determined by the fixed values of this feature coalition. The order of the remaining features is chosen as above to maximize comparability.
    \item[Randomize in random order] As a baseline, we randomize the features in random order to see how fast the prediction stability drops when there are no \enquote{special} features.
\end{description}

In order to obtain the interventional distributions required in \namenos, we follow the reasoning of \citet{janzing-2020} and consider the real world to be a confounder of the input features of the ML model, which therefore do not have direct causal links among each other. As a consequence, we can intervene directly on the input features by simply fixing the values of the features belonging to the considered coalition while sampling the remaining features from their (joint) marginal distribution.
Further, we choose a threshold of $\alpha = 0.9998$ to identify coalitions with (almost) full explanations while allowing for numerical errors. Additionally, we set the maximum coalition size to $k=4$ and stop the procedure if we do not find a coalition exceeding the threshold at this size. 

\subsubsection{Result}
Out of the $1,000$ samples in the credit dataset, $426$ have simple explanations as defined above (most of them have multiple such explanations, such that the mean and quantiles in \cref{fig:credit} are composed of 1,803 data points). On average, the cumulative normalized SHAP value of the minimum-size full explanations is $53.2\%$.

Analyzing \cref{fig:credit}, we see that, on average, \name and SHAP show a similar performance, although \namenos' performance for the solid-line sample (sample 500) only drops for the last 3 features while SHAP decreases a few fetures earlier. This is due to the fact that, for this sample, a 3-feature coalition is actually sufficient to reach an explanation score of $\approx 1$. Grouping the samples by the size $k$ of their minimum-size full explanation coalitions, we would see that the stability of \name is very much determined by this value, i.e., the prediction stability always drops after $|\bm{X}|-k$ features. Such a pattern cannot be observed for SHAP.

Using a random order for the randomization of features, the drop in the prediction stability starts much earlier, usually after $\approx 10$ features. Note that there is a small number of outliers even after one or two features, causing the mean to decrease immediately, while the 5-95\% range is still at 1.

\subsubsection{Computation}
The experiment was executed on a 10-core MacBook Pro M2. The calculation of the SHAP values took $\approx 5$ minutes, while the identification of all minimum-size coalitions up to size $k=4$ took $\approx 6$ hours.

\subsection{Identifying root causes in a cloud computing application} \label{app:cloud}
We consider a cloud computing application in which services call each other to produce a result which is returned to the user at the target service $Y$. Services can either be stand-alone (like a database or a random number generator) or call other services and transform their input to produce an output. When we observe an error at the target service $Y$, we want to know which service(s) caused this error. 

\subsubsection{Data generation}
\begin{figure}[h]
	\centering
	\includegraphics[width=0.8\textwidth]{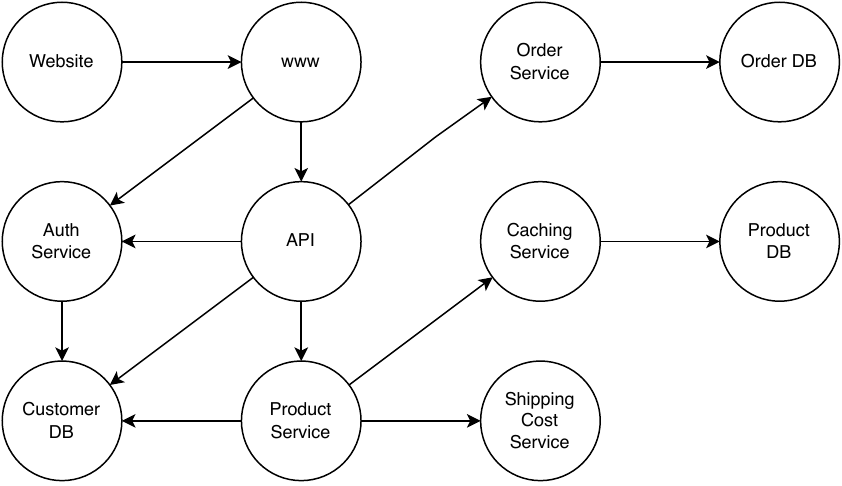}
	\caption{Cloud service architecture with non-trivial causal dependencies. \label{fig:cloud-services}}
\end{figure}

\begin{figure}[t]
	\centering
	\includegraphics[width=0.8\textwidth]{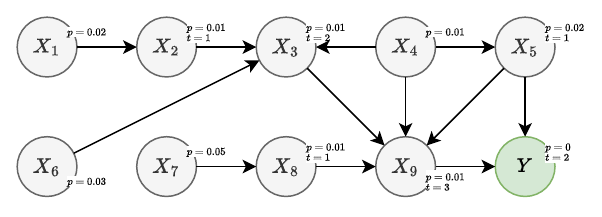}
	\caption{Causal graph of the cloud computing application in \cref{fig:cloud-services}. The parameters $p_i$ denote the probability that a node produces an error itself, and $t_i$ is the threshold number of parent services that have to encounter an error to also cause the error to propagate to $X_i$. \label{fig:cloud-network}}
\end{figure}

The causal network of the Cloud Computing Application experiment is based on a realistic service architecture shown in \cref{fig:cloud-services}. On top of the causal dependencies, we have defined parameters $p_i$ and $t_i$ such that target errors are a rare, but not impossible phenomenon.
This architecture translates to the causal network shown in \cref{fig:cloud-network} with services $X_1, ..., X_9, Y$, by omitting the \emph{Website} service and using \emph{www} as the target variable. We generate synthetic data representing error counts for the different services according to 
\begin{align}
X_i=\Lambda_i \vee \left(\sum_{j \in \parents{i}}{X_j} \geq t_i\right) \text{, where } \Lambda_i \sim \text{Bern}(p_i) \; .
\end{align}
We generated $10$ million samples, of which $15,373$ samples produce a target error (i.e., $Y=1$). \Cref{tab:cloud-number-of-samples} shows the number of elements in the sample set, grouped by the number of original errors. We see that fewer than two original errors never lead to a target error.

\begin{table}[t]
\centering
\caption{Number of generated samples, grouped by the number of original errors, i.e., noise nodes with value 1. \label{tab:cloud-number-of-samples}}

    \begin{tabular}{ccc}
        \toprule
        \# errors & \# samples with $Y=0$ & \# samples with $Y=1$ \\
        \midrule
        0 & 8,416,434 & 0 \\
        1 & 1,471,666 & 0 \\
        2 & 93,532 & 14,009 \\
        3 & 2,953 & 1,283 \\
        4 & 42 & 78 \\
        5 & 0 & 3 \\
        \bottomrule
    \end{tabular}
\end{table}

\subsubsection{Mapping from root causes to minimal coalitions}
From the contribution scores produced by the RCA methods \citep[see][]{budhathoki-2022}, we find minimal coalitions with the following procedure:
\begin{enumerate}
	\item Calculate the contribution score $\Phi(i)$ for all features $i \in I$.
	\item Sort the features by descending score and normalize them so that $\sum{\Phi_i}=1$.
  \item Select feature via a \emph{cumulative threshold} (take features in the above order until the sum of their contribution scores reaches a threshold $\theta_c=0.95$), or via an \emph{individual threshold} (take all features with $\Phi(i) \geq \theta_i=0.15$).
\end{enumerate}
The thresholds $\theta_c$ and $\theta_i$ were found through hyperparameter tuning such that these RCA methods achieve the best performance when compared to the ground truth.  
By applying this procedure, we get a coalition of features from all baseline methods.

\subsubsection{Traversal RCA} \label{app:cloud-traversal}
The traversal method for RCA is based on a very simple idea: A node in the causal graph is identified as a root cause if (a) it shows an error (we say the node is \emph{anomalous}), and (b) none of its causal parents is anomalous. This can be directly inferred from an observation, outputting a set of root causes. 

\begin{figure}[t]
	\centering
    \begin{subfigure}[t]{0.47\textwidth}
        \includegraphics[width=\textwidth]{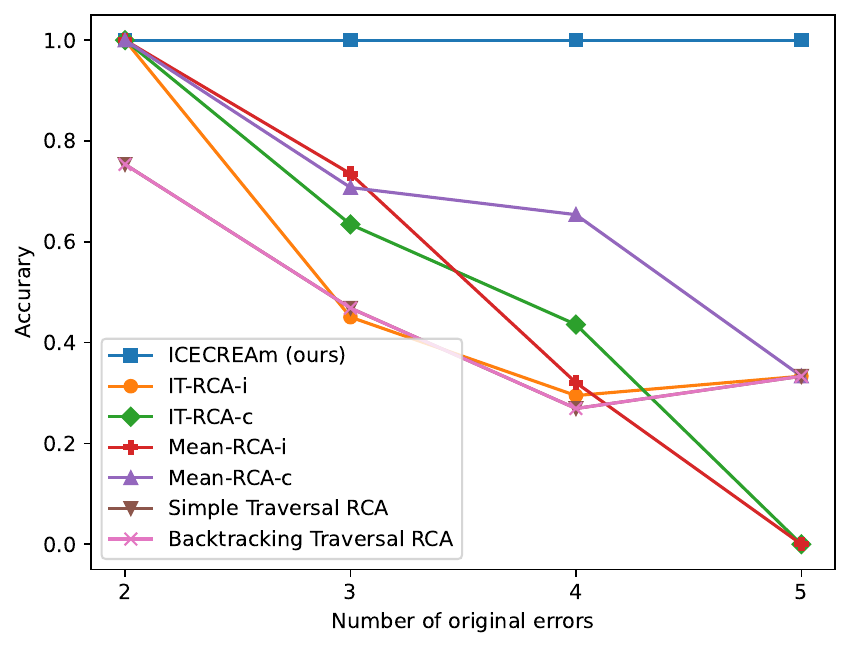}
        \caption{Accuracy of identifying root causes for different numbers of injected errors. \label{fig:cloud-traversal-accuracy}}    
    \end{subfigure}
    \hfill
    \begin{subfigure}[t]{0.47\textwidth}
        \includegraphics[width=\textwidth]{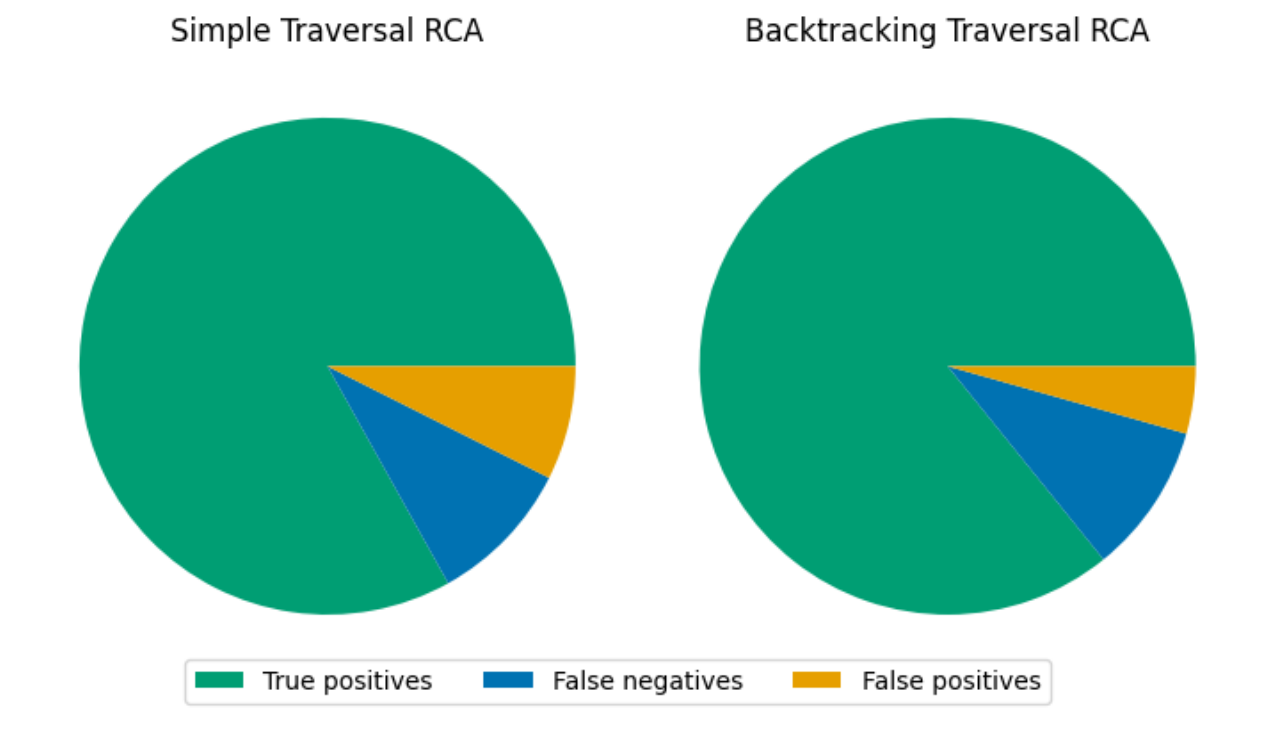}
         \caption{True positives, false positives, and false negatives for the cases with at least 3 injected errors. \label{fig:cloud-traversal-confusion}}
    \end{subfigure}	
	\caption{Comparison of Simple Traversal RCA and Backtracking Traversal RCA for the Cloud experiment. The accuracy is unaffected, while the false positive rate decreases when using backtracking.} \label{fig:cloud-traversal}
\end{figure}

A slightly more sophisticated way is to additionally require for a root cause to have a path of anomalous nodes to the target, therefore backtracking from the target to possible root causes (called \emph{Backtracking Traversal RCA} in \cref{fig:cloud-traversal}). This can be implemented as a breadth-first search from the target node which stops at non-error nodes and then returns the leaves of the search tree. The effect is that a node is only identified as a root cause if there is a direct error propagation path from this node to the target, thus reducing false positives.

For completeness, we show in \cref{fig:cloud-traversal} a comparison between these two traversal RCA methods. Although the false positive rate, as expected, is lower for Backtracking Traversal RCA, the accuracy remains the same: The main failure mode of Traversal RCA is the implicit assumption that a single anomalous parent is enough to prevent a node from being a root cause. This weakness is present in both versions of the algorithm, as witnessed by the high rate of false negatives.

\subsubsection{Computation}
Most of the experiment and analysis was executed on a 10-core MacBook Pro M2. Identifying all minimum-size coalitions for the 15,000 target error samples took approximately one hour; the traversal algorithms did not take any substantial amount of time ($<20$ seconds). However, calculating the anomaly scores for the RCA approaches using \texttt{dowhy.gcm} \citep{bloebaum-2022} turned out to be much slower, such that this task was transferred to 10 EC2 instances and finished after about two days.

\end{document}